\documentclass[letterpaper, 10 pt, conference]{ieeeconf}  

\IEEEoverridecommandlockouts                              
\pdfoutput=1

\usepackage{graphicx,amsmath,amssymb,bm,xcolor,nicefrac}
\usepackage[pass,a4paper]{geometry}
\usepackage{subfigure} 

\newcommand{\trsp}{{\scriptscriptstyle\top}}

\newcommand{\pos}{{\scriptscriptstyle\text{pos}}}
\newcommand{\ori}{{\scriptscriptstyle\text{ori}}}
\newcommand{\kin}{{\scriptscriptstyle\text{kin}}}

\usepackage[hidelinks]{hyperref}

\title{\LARGE \bf
Imitation of Manipulation Skills Using Multiple Geometries
}

\author{Boyang Ti$^{1,2}$, Yongsheng Gao$^{1}$, Jie Zhao$^{1}$ and Sylvain Calinon$^{2}$
\thanks{This paper is supported by the China Scholarship Council (CSC, No.202006120159), by the European Commission's Horizon 2020 Programme through
the MEMMO Project (Memory of Motion, https://www.memmo-project.eu/, grant agreement 780684) and by the National Natural Science Foundation of China (No. 92048301).}
\thanks{$^{1}$State Key Laboratory of Robotics and System, Harbin Institute of Technology, Harbin 150001, China (email: 17b908043@stu.hit.edu.cn; gaoys@hit.edu.cn; jzhao@hit.edu.cn)}
\thanks{$^{2}$Idiap Research Institute, CH-1920 Martigny, Switzerland (email: sylvain.calinon@idiap.ch)}%
}

\begin{document}

\maketitle
\thispagestyle{empty}
\pagestyle{empty}

\begin{abstract}
Daily manipulation tasks are characterized by geometric primitives related to actions and object shapes. Such geometric descriptors are poorly represented by only using Cartesian coordinate systems. In this paper, we propose a learning approach to extract the optimal representation from a dictionary of coordinate systems to encode an observed movement/behavior. This is achieved by using an extension of Gaussian distributions on Riemannian manifolds, which is used to analyse a set of user demonstrations statistically, by considering multiple geometries as candidate representations of the task. We formulate the reproduction problem as a general optimal control problem based on an iterative linear quadratic regulator (iLQR), where the Gaussian distribution in the extracted coordinate systems are used to define the cost function. We apply our approach to object grasping and box opening tasks in simulation and on a 7-axis Franka Emika robot. The results show that the robot can exploit several geometries to execute the manipulation task and generalize it to new situations, by maintaining the invariant characteristics of the task in the coordinate system(s) of interest.
\end{abstract}

\section{INTRODUCTION}

Skillful manipulation does not only relate to the precision that a person can achieve. More importantly, it relates to the capability of exploiting variations, allowing skilled performers to better take advantage of this redundancy to counteract perturbations that could impact the fulfillment of the task, while ignoring other perturbations. This finding stems from various disciplines, including biomechanics, neuroscience, sport science, control and robotics, with related formulations including minimal intervention principle \cite{Todorov02b}, uncontrolled manifold \cite{Scholz99} or optimal feedback control \cite{Wolpert11}. For example, in the uncontrolled manifold model, if the analysis of movements reveals that the variability in the \emph{do-not-matter} directions is larger than in the other orthogonal directions, this statistics is taken as an evidence of skill, as control is not exerted where it does not matter. Namely, the central nervous system focuses its control effort on variables that matter for the task, where the variability is not randomly scattered, but instead channeled preferentially along the \emph{do-not-matter} directions \cite{Sternad10}. 

The importance of variations to model and evaluate skillful manipulations also extends to correlations, including the well-known importance of coordination in movements. This role can even be extended to the more general principle of synergies as the main principle used by nature to handle complexity in biological systems \cite{Kelso09}. In this formulation, a perturbation to any part of the synergy is immediately compensated for by remotely linked elements in such a way as to preserve the functional integrity or the goals of the organism.

The modeling of (co)variations is tightly linked to the problem of defining coordinate systems in which this analysis takes place \cite{Sternad10}. The achievement of a manipulation task can be characterized by multiple options. Although these different solutions achieve the same end goal, all solutions are not equivalent as they may differ in the way they forgive errors. In other words, the acquisition of skills not only decreases variability but also takes advantage of the structure of the task.

In motion planning, the advantage of using different coordinate systems or even hybrid coordinate systems has been demonstrated from geometrical and computational perspectives, see \cite{Laumond18} for an overview. For human movement planning, evidence has accumulated that the parietal cortex codes for movement in the head- or gaze-centered coordinate frames, the putamen in a body reference frame, and the hippocampus in an environmental reference frame, also suggesting a neural basis for the use of several geometries \cite{Bennequin09,Ganesh12}.

\begin{figure}[t]
	\centering
	\includegraphics[width=\columnwidth]{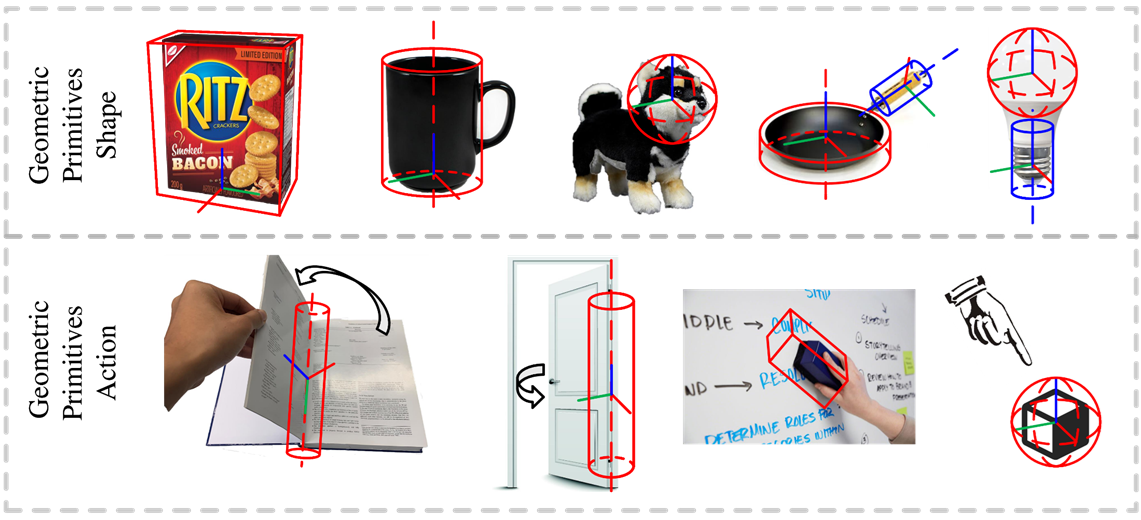}	
	\caption{Geometric primitives in daily objects and tasks, with examples involving prismatic, cylindrical and spherical shapes.}
	\label{Fig6}
	\vspace{-0.3cm}
\end{figure}

In this paper, we present a motion planning approach for robot manipulation that allows the most relevant coordinate systems to be extracted by statistics. This problem is formulated as a general optimal control problem, where Gaussian distributions on Riemannian manifolds are used to define a cost function, whose parameters are learned from a small set of demonstrations.

The Riemannian geometry formulation allows a wide range of geometries to be taken into account. We demonstrate here the potential of the approach by considering three candidate geometries that seem to be particularly relevant for structured human-made environments, which correspond to prismatic, cylindrical and spherical coordinate systems. An iterative linear quadratic regulator (iLQR) approach is proposed by considering several coordinate systems in the underlying cost function as candidates to describe a manipulation skill, whose precision requirements are learned from the regularities in the observed demonstrations. The proposed approach thus automatically selects an appropriate metric to describe the task, where iLQR is exploited to generalize the task to new situations.

The consideration of this specific subset of Riemannian manifolds is driven by the observation that most common objects are composed of geometric primitive shapes, either representing the complete object or a local part of the object that is relevant at a given phase of the task for grasping and manipulation, as shown in the first row of \autoref{Fig6}. Moreover, manipulation of objects in a structured environment can be expressed with constraints by using similar geometries. For example, displacing objects that rotate along an axis, such as opening a door, turning the pages of a book, or tilting furniture, are all ideally described within a cylindrical coordinate system, independently of the original shape of the manipulated object, as shown in the second row of \autoref{Fig6}. Indeed, the overall shape of a book is prismatic, but the action of pointing toward this book has a more straightforward description in a spherical coordinate system, while the action of turning its pages has a simpler description in a cylindrical coordinate system (with axis oriented along its binding). Notably, the variations characterizing these actions are easier to describe within these specific coordinate systems, which also help at rejecting perturbations in a coordinated manner during the reproduction of the skill.

The contributions of our research are threefold: ($i$) we present an approach to improve the generalization capability of manipulation skills by considering different types of coordinate systems; ($ii$) we propose to construct a Gaussian distribution in multiple coordinate systems to represent invariant features of observed manipulation skills; ($iii$) we propose an approach to extract optimal coordinate systems and reproduce the learned skill with an optimal control strategy.

\section{Related Work}
Most of the objects in our daily life can be modeled with geometric primitives. In \cite{kaiser2019}, Kaiser \emph{et al.} showed that simple geometric primitives can provide compact and robust representation of objects. In \cite{romanengo2021,li2019}, geometric primitives are fitted to point clouds to reconstruct their shapes.

The consideration of geometry in manipulation skills has been investigated early on in robotics, including the pioneer work of Mason formalizing force control based on task geometry \cite{Mason81}. In most applications, three priority levels of safety, primary and auxiliary constraints constructed from task geometric primitives suffice to describe manipulation features \cite{borghesan2015}. 

In learning from demonstration (LfD), the consideration of geometry has been introduced in various ways \cite{Billard16chapter,Calinon19entry}. In \cite{PerezDArpino17,Subramani18}, demonstrations are used to extract geometric constraints from the observed regularities, by reasoning about Special Euclidean Group in 3 dimensions (SE(3)) volumetric and computer-aided design (CAD) constraints \cite{PerezDArpino17}, or from a list of kinematic constraints such as fixed point, axial rotation, prismatic motion or planar motion \cite{Subramani18}.

Another related approach consists of exploiting the observed regularities to extract soft constraints, in the form of full precision matrices information used to derive a controller with adaptive tracking gains or stiffnesses. This can for example be achieved by using the inverse of covariance matrices (corresponding to precision matrices) in the cost function of an optimal control problem, so that invariant task features are repeated with greater precision than task features allowing variations \cite{Calinon14ICRA}.

The optimal control problem formulation can also be used to solve planning problems, by considering time windows covering the entire task. The iterative linear quadratic regulator (iLQR) \cite{Mayne66,Li04} is of particular interest to our work, as it can typically be used to search for a solution in the joint angle configuration space of the robot, either in the form of open-loop control commands for simple planning, or as a full controller with feedback and feedforward terms for control. The approach allows the task to be specified as a sparse set of viapoints describing the position and orientation of the robot end-effector, which can be solved efficiently as an optimization problem (a Gauss--Newton iteration scheme in the planning case).

Standard iLQR problems are formulated with a cost expressed in a Cartesian coordinate system centered at the robot base or end-effector. In \cite{Calinon16JIST}, we demonstrated the advantage of considering several Cartesian coordinate systems centered on objects or landmarks of interest within a LQR control strategy expressed in task space, which required an inverse kinematics solution to be implemented as a separated process. We propose here to extend the approach to iLQR, which can solve the two problems in a unified manner. Moreover, we propose to extend the approach to coordinate systems that are not only attached to different objects or landmarks, but that also consider different geometries. 

Standard approaches to manipulation skill learning represent the target as a point to reach, which loses important shape information for adapting to new situations and perturbations. The advantage of our approach is to take the geometry information into account within an iLQR problem formulation, with a cost automatically specified based on the demonstrations. In a standard Cartesian coordinate system, the distributions corresponding to different grasping points (e.g., spread along a circular shape) cannot be well represented with a single Gaussian distribution. Fitting more complex distributions (e.g., a mixture of Gaussians) typically require numerous demonstrations to have a good representation of the distribution. In contrast, the use of multiple coordinate systems, as we propose in this work, limits the possible distributions to a subset of shapes, yielding an approach to learn which cost function to use in an iLQR problem from a very small set of demonstrations. Indeed, in the proposed approach, only a few points are needed in each coordinate system to estimate a single Gaussian for each phase of the movement. This approach is also consistent with the notion of shape descriptors used in grasping, which often form geometric primitives such as spherical, cylindrical and prismatic descriptors \cite{Eppner13}. We formalize the statistical analysis from the perspective of these different heterogeneous shapes by exploiting Gaussian distributions on Riemannian manifolds, see \cite{Calinon20RAM} for a review.

\section{Proposed approach}
In optimal control, a cost function is minimized with respect to control commands over a time window, with a dynamics function describing the evolution of the system by starting from an initial state. When the cost is composed of quadratic error terms on the states and control commands and the dynamics is linear, we have an LQR problem that can be solved either in batch or recursive form. For other costs and/or dynamics, the problem can be solved iteratively by starting from an initial estimate and by computing a first order Taylor expansion of the cost and dynamics, so that the problem can be solved as a Newton optimization problem (for open loop control commands), or as a Gauss--Newton optimization problem when the cost can be expressed with residuals in a quadratic form. The resulting approach is called iLQR \cite{Mayne66,Li04}, where each step of the optimization solves an LQR problem locally, either with batch (open loop controller) or recursive formulation (feedback controller). 

We use the batch form in our experiment, by considering joint angle velocity commands, with the dynamics expressed as a single integrator. For a robot with $D$ articulations, by concatenating the joint angle states $\bm{q}_t$ and joint angle velocity commands $\bm{u}_t$ at different time steps $t\in\{1,\ldots,T\}$ as large vectors $\bm{q}$ and $\bm{u}$ of lengths $DT$, the evolution of the system is linear, expressed as $\bm{q}= \bm{S}_{\bm{q}} \bm{q}_0 + \bm{S}_{\bm{u}} \bm{u}$, where $\bm{q}_0$ is the initial joint state, and $\bm{S}_{\bm{q}}$ and $\bm{S}_{\bm{u}}$ are matrices describing the evolution of the system, see \cite{Lembono21IROS} for details. We define $\bm{x}_{n,t}$ as a vector composed of elements $x_{d,n,t}$, where $d$ represents the dimension of the manifold to represent the end-effector position at each timestep in different manifolds. We then define $\bm{x}_{n,t}=\bm{f}^\kin_{n,t}(\bm{q}_t)$ as the forward kinematics function at time step $t$ expressed in the coordinate system $n$.

We construct Gaussian distributions for each candidate manifold $n\in\{1,\ldots,N\}$ from a small set of demonstrations. For the computation of distances on a manifold $\mathcal{M}=\mathbb{R}^d$ of $d$ dimensions, corresponding to standard Cartesian spaces, and on a manifold $\mathcal{M}=\mathcal{S}^d$ of $d$ dimensions, corresponding to (hyper)spheres, we require logarithmic map functions (residuals), which can be computed analytically with
\textcolor{black}{
\begin{align}
	\text{Log}^{\mathbb{R}^d}_{\bm{\mu}}\!(\bm{x}) &= \bm{x} - \bm{\mu}, \\
	\text{Log}^{\mathcal{S}^d}_{\bm{\mu}}\!(\bm{x}) &= \arccos(\bm{\mu}^\trsp \bm{x}) \, \frac{\bm{x} - \bm{\mu}^\trsp \bm{x} \, \bm{\mu}}{\|\bm{x} - \bm{\mu}^\trsp \bm{x} \, \bm{\mu}\|}.
\end{align}
}

For each point $\bm{p}\!\in\!\mathcal{M}$, there exists a tangent space $\mathcal{T}_{\bm{p}} \mathcal{M}$ that locally linearizes the manifold. We use a simple approach that consists of estimating the mean of the Gaussian as a centroid on the manifold, and representing the dispersion of the data as a covariance expressed in the tangent space of the mean \cite{Pennec06,SimoSerra16,Zeestraten17RAL}. 
Such a distribution is defined as
\begin{equation}
	\mathcal{N}_{\mathcal{M}}(\bm{x}|\bm{\mu},\bm{\Sigma}) = 
	{\Big((2\pi)^{d} | \bm{\Sigma} |\Big)}^{-\frac{1}{2}} \; 
	e^{-\frac{1}{2} \text{Log}_{\bm{\mu}}\!(\bm{x}) \, \bm{\Sigma}^{-1} \, \text{Log}_{\bm{\mu}}\!(\bm{x})},
\end{equation}
where $\bm{x}\!\in\!\mathcal{M}$ is a point of the manifold, $\bm{\mu}\!\in\!\mathcal{M}$ is the mean of the distribution (origin of the tangent space), and $\bm{\Sigma}\in\mathcal{T}_{\bm{\mu}}\mathcal{M}$ is the covariance matrix defined in this tangent space.

For a set of $N$ datapoints, this geometric mean corresponds to the minimization
\begin{equation}
	\min_{\bm{\mu}} \sum_{n=1}^N {\text{Log}_{\bm{\mu}}\!(\bm{x}_n)}^\trsp \; 
	\text{Log}_{\bm{\mu}}\!(\bm{x}_n),
\end{equation}
which can be solved by a simple and fast Gauss--Newton iterative algorithm. This is done by iterating
\begin{equation}
	\bm{u} = \frac{1}{N} \sum_{n=1}^N \text{Log}_{\bm{\mu}}\!(\bm{x}_n),
	\qquad
	\bm{\mu} \leftarrow \text{Exp}_{\bm{\mu}}\!(\bm{u}),
\end{equation}
until convergence. After convergence, a covariance matrix is computed in the tangent space as $\bm{\Sigma}=\frac{1}{N} \sum_{n=1}^N \text{Log}_{\bm{\mu}}\!(\bm{x}_n) \, \text{Log}_{\bm{\mu}}^\trsp\!(\bm{x}_n)$, see \cite{Calinon20RAM} for details.

After evaluation of these Gaussians in different coordinate systems, we consider for each time step the coordinate system $\hat{n}$ that shows the most regularities between the different demonstrations (\emph{winner-takes-all} strategy), namely
\begin{equation}
	\hat{n} = \arg\min_n |\bm{\Sigma}_{n,t}|, 
	\quad\forall n \in\{1,\ldots,N\},
\end{equation}
where $|\bm{\Sigma}_{n,t}|$ is the determinant of the covariance matrix $\bm{\Sigma}_{n,t}$, and $N$ is the number of coordinate systems. 

We then define $\bm{f}_{\hat{n}}(\bm{q})$ as a concatenation of residual functions $\bm{f}_{\hat{n},t}(\bm{q}_t)$ at time steps $t\in\{1,\ldots,T\}$, selected from the set of residuals $\bm{f}_{n,t}(\bm{q}_{t})$ expressed in different coordinate systems, with $n=\{1,\ldots,N\}$.  The Jacobian matrix of $\bm{f}_{\hat{n}}(\bm{q})$ is $\bm{J}_{\hat{n}}(\bm{q})$. Similarly, the precision matrix $\bm{Q}_{\hat{n}}$ is a block-diagonal concatenation of precision matrices $\bm{Q}_{\hat{n},t}$ in the selected coordinate systems, expressed in the corresponding tangent spaces. The precision matrices $\bm{Q}_{\hat{n},t}$ are computed from covariance matrices as $\bm{Q}_{\hat{n},t}= \bm{\Sigma}_{\hat{n},t}^{-1}$.

We then define a cost function at each time step $t$ as
\begin{equation}
	c(\bm{q}_t,\bm{u}_t) = 
	{\big\| \bm{f}_{\hat{n},t}(\bm{q}_t) \big\|}^2_{\bm{Q}_{\hat{n},t}}  
	+ {\big\| \bm{u}_t \big\|}^2_{\bm{R}_t},
	\label{eq:cost}
\end{equation}
where the residual vector $\bm{f}_{\hat{n},t}(\bm{q}_t)=\text{Log}^{\mathcal{M}_{\hat{n}}}_{\bm{\mu}_{\hat{n},t}}\!(\bm{x}_{\hat{n},t})$ is expressed in the tangent space of $\bm{\mu}_{\hat{n},t}$. $\bm{R}_t$ is a control weight matrix. The logarithmic map $\text{Log}(\cdot)$ is used to compute the first part of the cost \eqref{eq:cost} as a geodesic distance between $\bm{x}_{\hat{n},t}$ and $\bm{\mu}_{\hat{n},t}$ on the manifold $\mathcal{M}_{\hat{n}}$, weighted by the full precision matrix $\bm{Q}_{\hat{n},t}$.

According to the above cost function, by concatenating the above vectors and matrices for all timesteps $t$, each iLQR step computes the Gauss--Newton update
\begin{multline}
	\Delta\bm{\hat{u}} \!=\! {\Big(
		\bm{S}_{\bm{u}}^\trsp \bm{J}_{\hat{n}}(\bm{q})^\trsp \bm{Q}_{\hat{n}} \bm{J}_{\hat{n}}(\bm{q}) \bm{S}_{\bm{u}} \!+\! \bm{R}\Big)}^{-1} \\
	\Big(- \bm{S}_{\bm{u}}^\trsp \bm{J}_{\hat{n}}(\bm{q})^\trsp \bm{Q}_{\hat{n}} \bm{f}_{\hat{n}}(\bm{q}) - \bm{R} \, \bm{u} \Big).
	\label{eq:duh}
\end{multline}
where $\bm{J}_{\hat{n}}(\bm{q})$ is the Jacobian of $\bm{f}_{\hat{n}}(\bm{q})$, which is used with a backtracking line search strategy, see \cite{Lembono21IROS} for details of the overall iLQR process. 




In our experiments, we consider full end-effector poses with $\bm{x}_{\hat{n},t}={[{\bm{x}^\pos_{\hat{n},t}}^\trsp, {\bm{x}^\ori_{\hat{n},t}}^\trsp]}^\trsp$ and $\bm{\mu}_{\hat{n},t}={[{\bm{\mu}^\pos_{\hat{n},t}}^\trsp, {\bm{\mu}^\ori_{\hat{n},t}}^\trsp]}^\trsp$, so that each manifold is composed of a position part and an orientation part with $\mathcal{M}_{\hat{n}}=\mathcal{M}^\pos_{\hat{n}}\times\mathcal{M}^\ori_{\hat{n}}$, which exploits the Cartesian product property of Riemannian geometry.

\vspace{2mm}\noindent In 2D space, we consider $N\!=\!2$ coordinate systems:
\begin{enumerate} 
	\item Cartesian: $\mathcal{M}^\pos_1=\mathbb{R}^2$, $\mathcal{M}^\ori_1=\mathcal{S}^1$;
	\item Polar: $\mathcal{M}^\pos_2=\mathcal{S}^1\times\mathbb{R}^1$ (corresponding to polar angle and radius), $\mathcal{M}^\ori_2=\mathcal{S}^1$.
\end{enumerate} 

\noindent In 3D space, we consider $N\!=\!3$ coordinate systems:
\begin{enumerate} 
	\item Cartesian: $\mathcal{M}^\pos_1=\mathbb{R}^3$, $\mathcal{M}^\ori_1=\mathcal{S}^3$;
	\item Cylindrical: $\mathcal{M}^\pos_2=\mathcal{S}^1\times\mathbb{R}^2$ (corresponding to polar angle, radius and height), $\mathcal{M}^\ori_2=\mathcal{S}^3$;
	\item Spherical: $\mathcal{M}^\pos_3=\mathcal{S}^2\times\mathbb{R}^1$ (corresponding to polar angle, azimuth angle and radius), $\mathcal{M}^\ori_3=\mathcal{S}^3$.
\end{enumerate}
$\mathcal{S}^d$ and $\mathbb{R}^{d}$ represent sphere and Cartesian manifolds of dimension $d$, respectively.

All the distances in the cost function \eqref{eq:cost} can be computed by exploiting the Cartesian product property of Riemannian manifolds, providing the composition rule
\begin{align} 
	\text{Log}^{\mathcal{M}_a\times\mathcal{M}_b}_{\bm{\mu}}\!(\bm{x}) &= 
	\begin{bmatrix} \text{Log}^{\mathcal{M}_a}_{\bm{\mu}_a}\!(\bm{x}_a) \\ \text{Log}^{\mathcal{M}_b}_{\bm{\mu}_b}\!(\bm{x}_b) \end{bmatrix}.	
\end{align}

\begin{figure}
	\centering
	\includegraphics[width=\columnwidth]{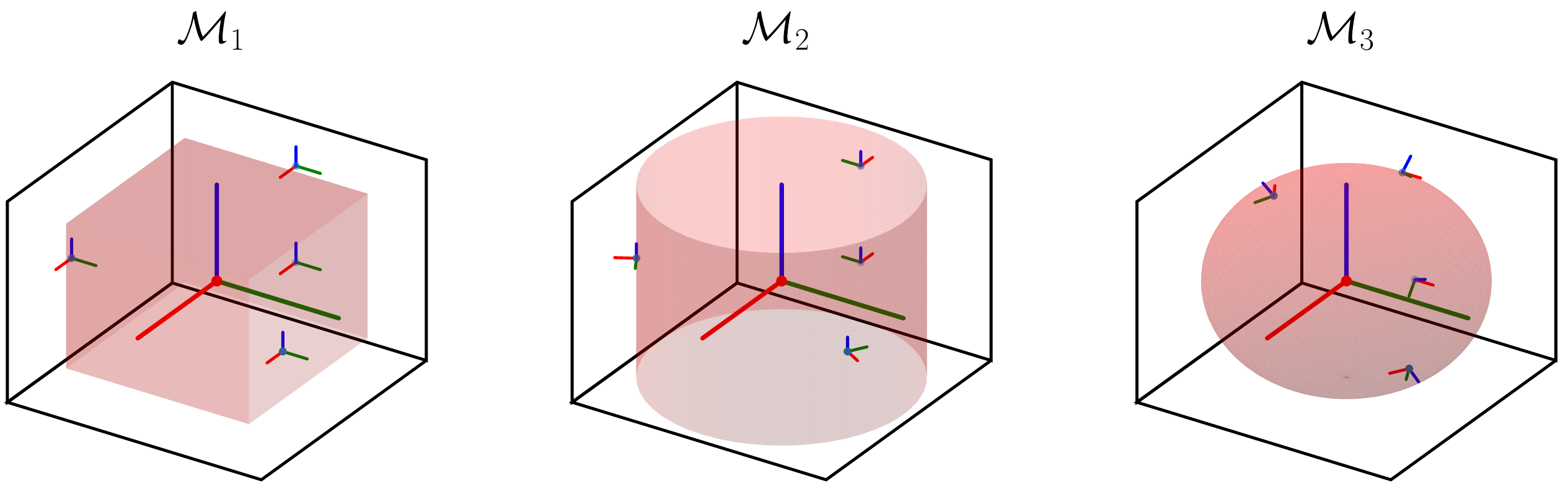}	
	\caption{Base frames of different manifolds $\mathcal{M}_1$ (i.e., $\mathbb{R}^3$), $\mathcal{M}_2$ (i.e., $\mathbb{R}^2 \times \mathcal{S}^1$) and $\mathcal{M}_3$ (i.e., $\mathbb{R}^1 \times \mathcal{S}^2$)}.
	\label{Fig8}
	\vspace{-0.3cm}
\end{figure}

For manifolds $\mathbb{R}^d$ and $\mathcal{S}^d$ of $d$ dimensions, the logarithmic maps can be computed analytically.
The orientation of the end-effector is expressed in a base frame following the shape of the manifold, by parallel transport of a canonical basis $\bm{I}$ defined at the origin of the manifold, see \autoref{Fig8}.


\section{Experiments}
We considered grasping and box opening as two typical manipulation tasks in our daily life to validate our approach. We evaluated the approach with a 3-DoF simulated planar robot and with a real 7-DoF Franka Emika manipulator.

\subsection{Grasping simulation}
\begin{figure}
	\centering
	\subfigure[Positions of the end-effector w.r.t.\ the object to grasp]
	{
		\includegraphics[width=0.95\columnwidth]{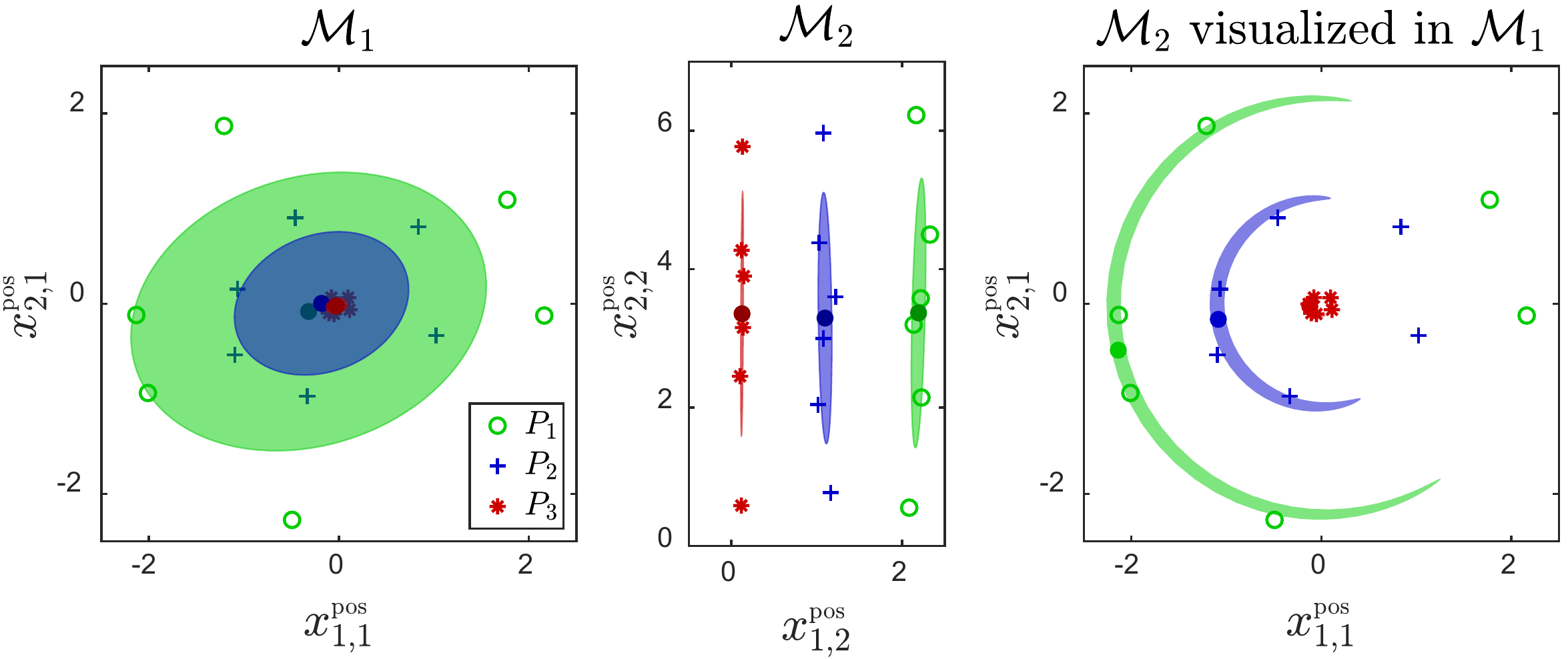}	
		\label{Fig1_a}
	}
	\subfigure[Orientations of the end-effector w.r.t.\ the object to grasp]
	{
		\includegraphics[width=0.95\columnwidth]{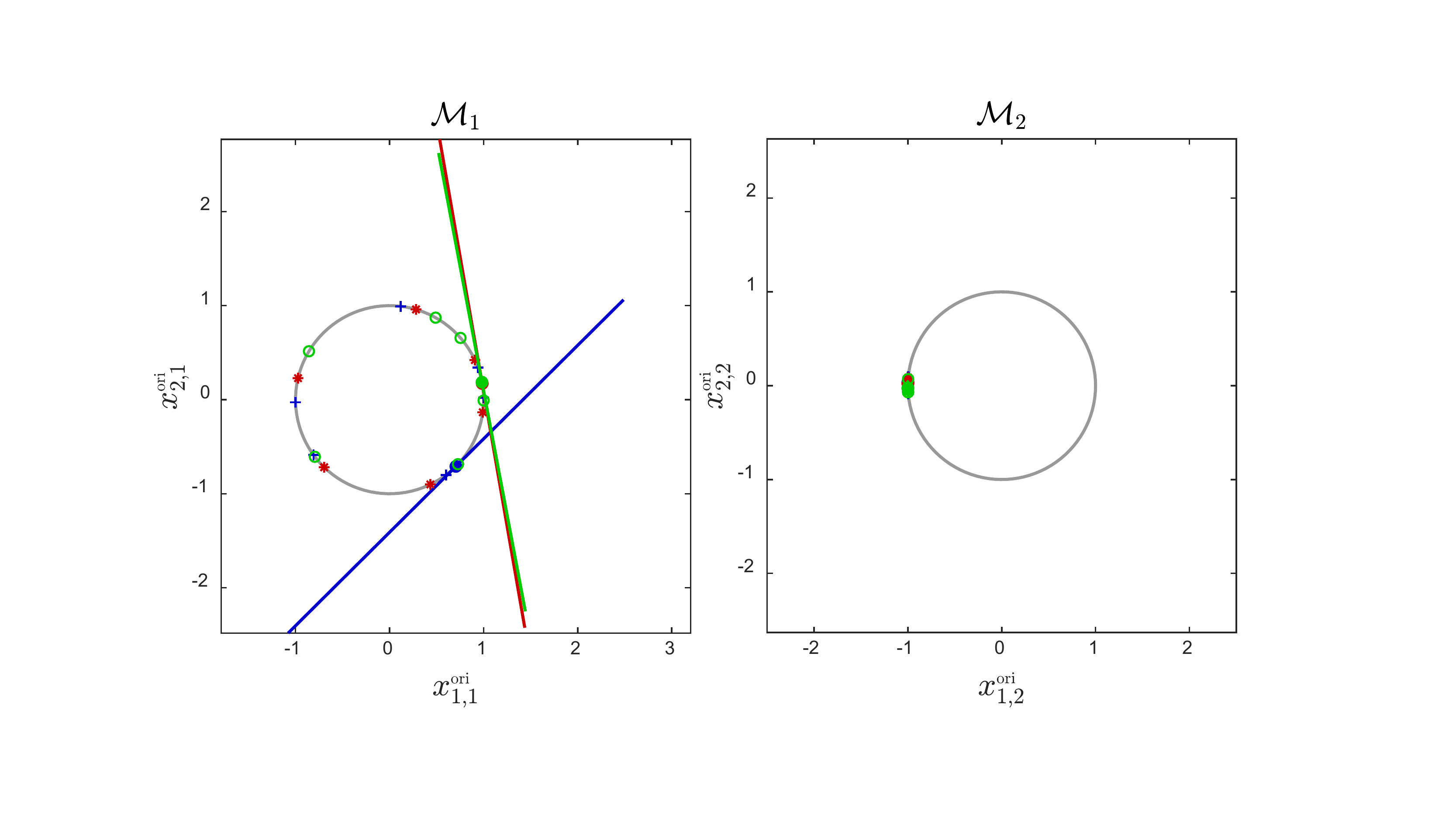}	
		\label{Fig1_b}
	}		
	\caption{Distributions of the three sets of stepwise reference ($\circ,\texttt{+},\star$) in $\mathcal{M}_1$ and $\mathcal{M}_2$. The indices in the graphs correspond to $x_{d, n}$. (a) The position is represented in two coordinate systems. The ellipses represent the Gaussian distributions for the three different phases of the motion (contours of one standard deviation). The contours of the Gaussian in $\mathcal{M}_2$ is also visualized in $\mathcal{M}_1$ to show that this distribution is a better fit (compare the left and right graphs). (b) The orientation is represented in the base frame of the two coordinate systems. The solid lines represent the variance in the tangent space of the $\mathcal{S}_1$ manifold, and the points represent the mean on the $\mathcal{S}_1$ manifold.}
	\label{Fig1}
	\vspace{-0.3cm}
\end{figure}

We generated six sets of demonstrations and divided each movement into three phases, which were used to construct Gaussian distributions as stepwise references in $\mathcal{M}_1$ and $\mathcal{M}_2$, as shown in \autoref{Fig1}. In $\mathcal{M}_1$, we observe from the left plot of \autoref{Fig1_a} that the means of positions almost overlap, with large covariances ignoring the circular organization of the datapoints. Similarly, in the left plot of \autoref{Fig1_b}, we see large variations in orientations when expressed in $\mathcal{M}_1$. In contrast, for the representation in $\mathcal{M}_2$, as shown in the middle plot of \autoref{Fig1_a}, the positions for each phase of the movement are well separated with low variance along $\bm{x}_{1,2}^\pos$, which encodes well the decrease of radius in each phase of the movement when approaching the object. The high variance along $\bm{x}_{2,2}^\pos$ reveals that the object can be approached from different directions. Similarly, in the right plot of \autoref{Fig1_b}, we see a low variance in orientation, reflecting the importance of orienting the gripper toward the object, in order to anticipate the grasping of this object. The determinants of the covariance matrices in the two coordinate systems in different phases are reported in \autoref{table2}, which are used to select a winning coordinate system at each time step of the motion. The system finds here that $\mathcal{M}_2$ is the best manifold to express the grasping motion.

\begin{figure}
	\centering
	\includegraphics[width=\columnwidth]{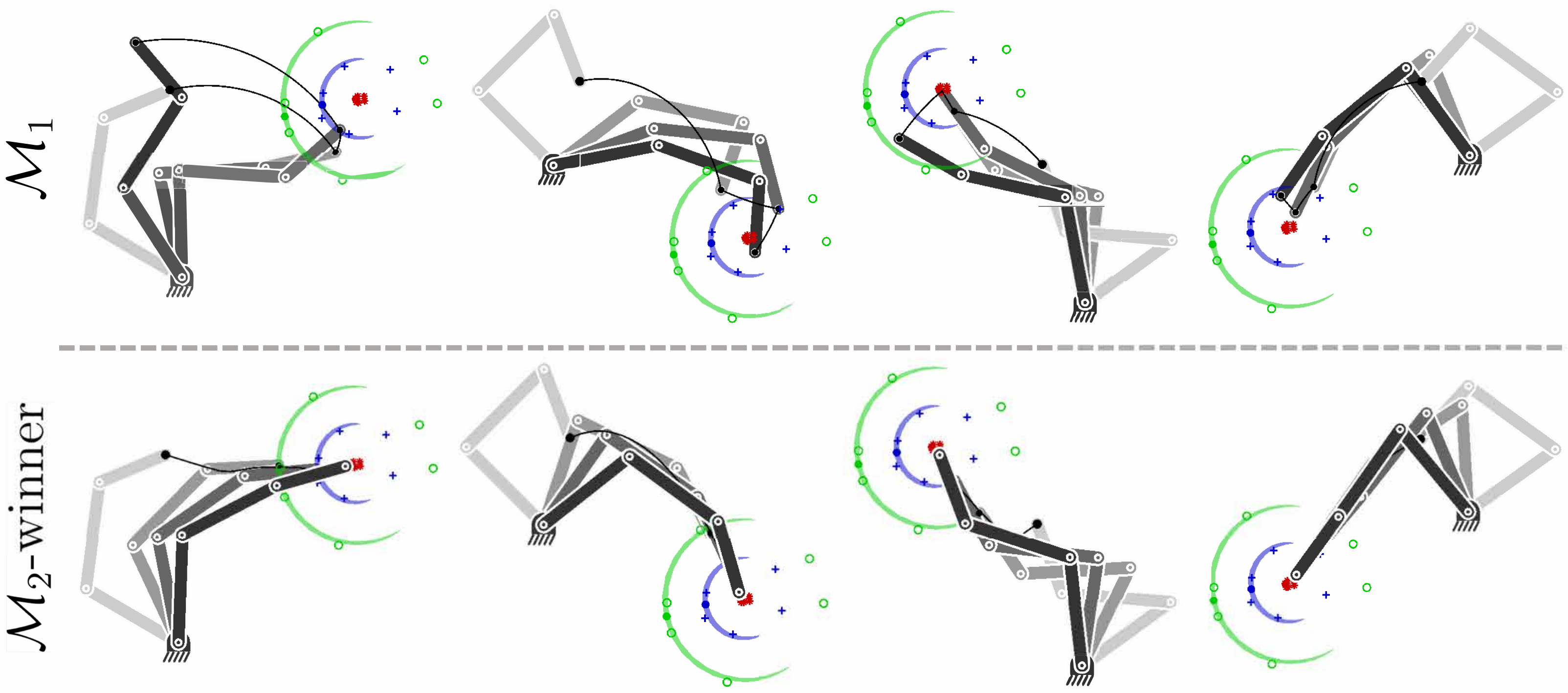}	
	\caption{The grasping movement is simulated in $\mathcal{M}_1$ and $\mathcal{M}_2$ (winner) from four different initial states of the robot (generated randomly). The movements of the robot are depicted with different shades of gray, with the lightest depicting the initial state of the robot.}
	\label{Fig2}
\end{figure}

To evaluate the generalization capability of our approach, we substitute the Gaussian distributions constructed in $\mathcal{M}_2$ (winner) and $\mathcal{M}_1$ into the cost function of the iLQR framework, with a random initial robot state. The desired movement should approach the object while pointing to it. The generalized movements are shown in \autoref{Fig2}. We observe that with $\mathcal{M}_1$, the motion cannot reproduce the movement well. In contrast, with $\mathcal{M}_2$ (the coordinate system selected by the algorithm), the robot keeps pointing toward the object while approaching it as expected, by correctly exploiting the variations allowed by the task.

\begin{figure}
	\centering
	\subfigure[Positions of the end-effector w.r.t.\ the box to open]
	{
		\includegraphics[width=0.95\columnwidth]{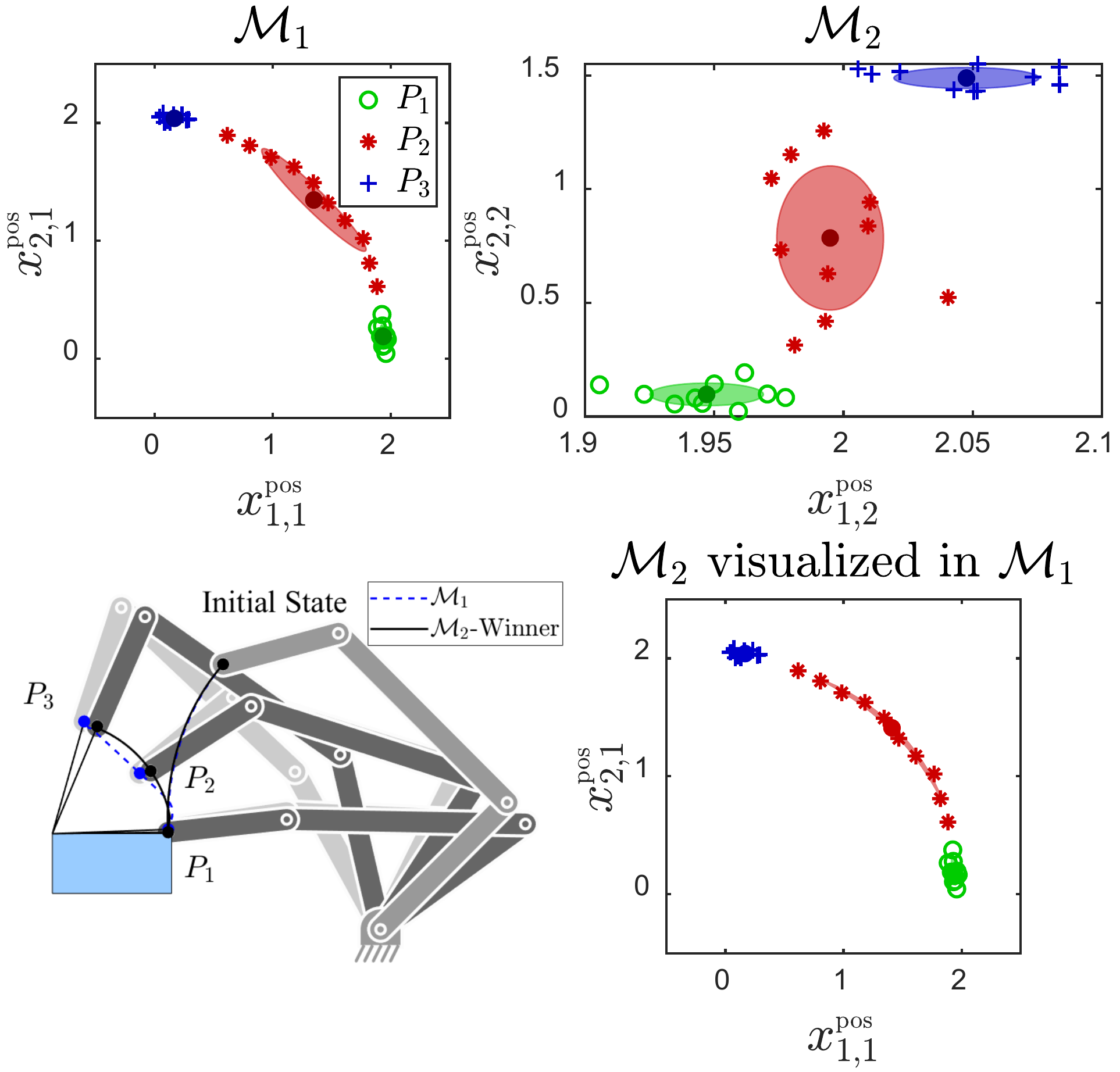}	
		\label{Fig3_a}
	}
	\subfigure[Orientations of the end-effector w.r.t.\ the box to open]
	{
		\includegraphics[width=0.95\columnwidth]{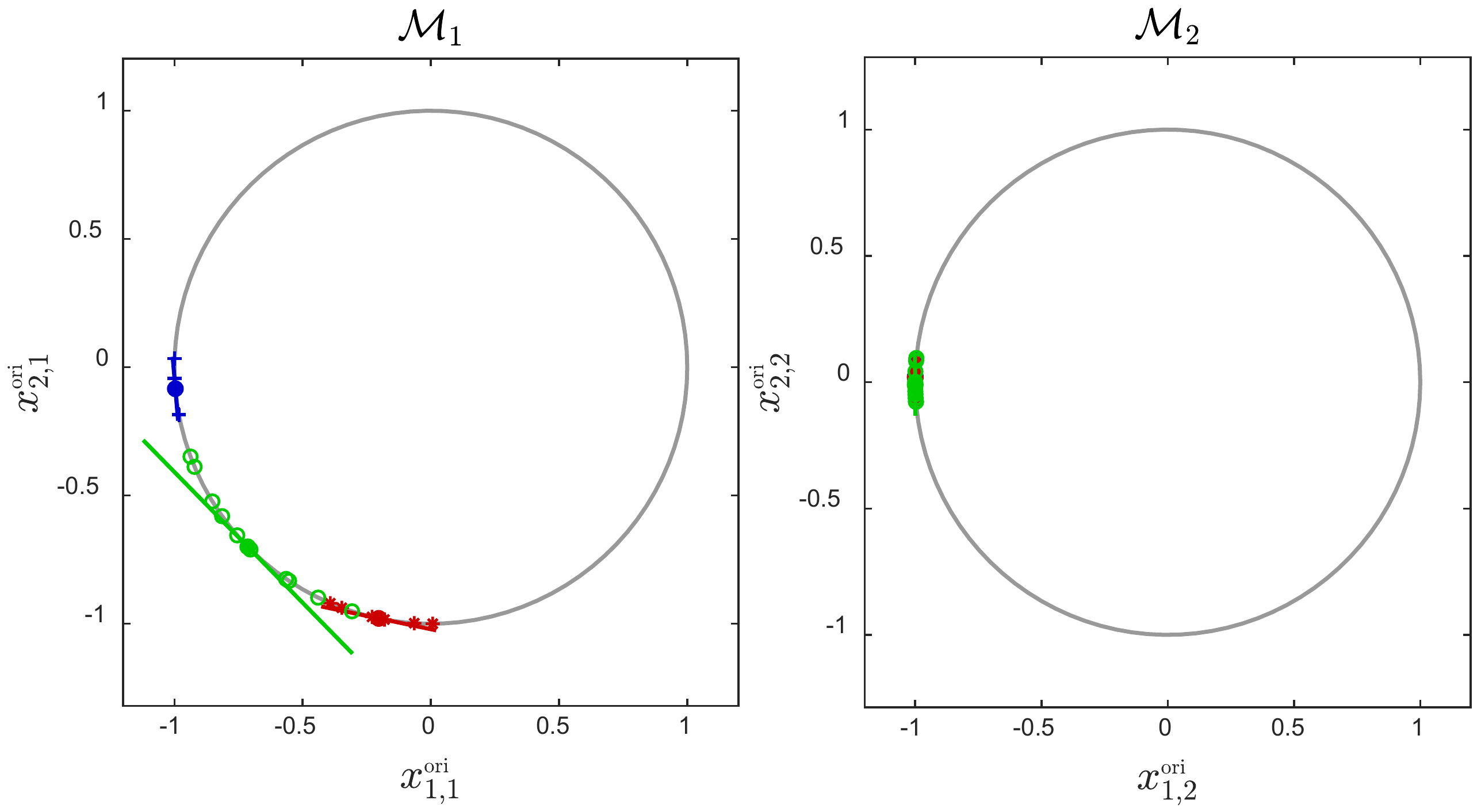}	
		\label{Fig3_b}
	}
	\caption{Box opening simulation. The indices in the graphs correspond to $x_{d, n}$. (a) Gaussian distributions for positions in the three phases of the motion, expressed in $\mathcal{M}_1$ and $\mathcal{M}_2$. The trajectories generated by using these two options are shown in the bottom-left plot, where $P_1$, $P_2$, $P_3$ represent the three phases. (b) Gaussian distributions for orientations in the three phases of the motion, expressed in $\mathcal{M}_1$ and $\mathcal{M}_2$.}
	\vspace{-0.3cm}
\end{figure}

\subsection{Box opening simulation}
We generated one demonstration and divided it into three phases, which were used to construct Gaussian distributions as stepwise references in $\mathcal{M}_1$ and $\mathcal{M}_2$. \autoref{Fig3_a} shows different distributions for $\mathcal{M}_1$ and $\mathcal{M}_2$. The algorithm correctly selects $\mathcal{M}_2$ as the most appropriate coordinate system throughout the movement, by correctly encoding the radius should be maintained for a skillful opening of the box, which is reflected by the small variance of $x_{1,2}^\pos$, see also \autoref{table2}. The bottom-left plot of \autoref{Fig3_a} shows that the regenerated movement in $\mathcal{M}_2$ maintains the circular shape of the action, which is not the case for $\mathcal{M}_1$. 

\begin{table}
	\renewcommand{\arraystretch}{1.2} 
	\setlength{\tabcolsep}{8pt}
	\centering 
	\caption{Determinants of covariance matrices at different phases of the motion for the grasping and box opening simulations.}
	\begin{tabular}{cccc}
		\hline
		\bfseries  & \multicolumn{3}{c}{Grasping / Box opening} \\
		\hline
		&$P_1$&$P_2$&$P_3$\\
		$\mathcal{M}_1$& {3.2e1 / 4.3e-6} & {2.2e0 / 3.0e-3} & {3.5e-4 / 6.8e-6}\\
		$\mathcal{M}_2$ & {\textbf{3.9e-5} / \textbf{1.1e-6}} & {\textbf{2.0e-5} / \textbf{8.2e-8}} & {\textbf{1.2e-6} / \textbf{1.9e-6}}\\
		\hline
	\end{tabular} 
	\label{table2} 
	\vspace{-0.3cm}
\end{table}

\subsection{Grasping tasks with the Franka Emika robot}
\begin{figure}
	\centering
	\includegraphics[width=\columnwidth]{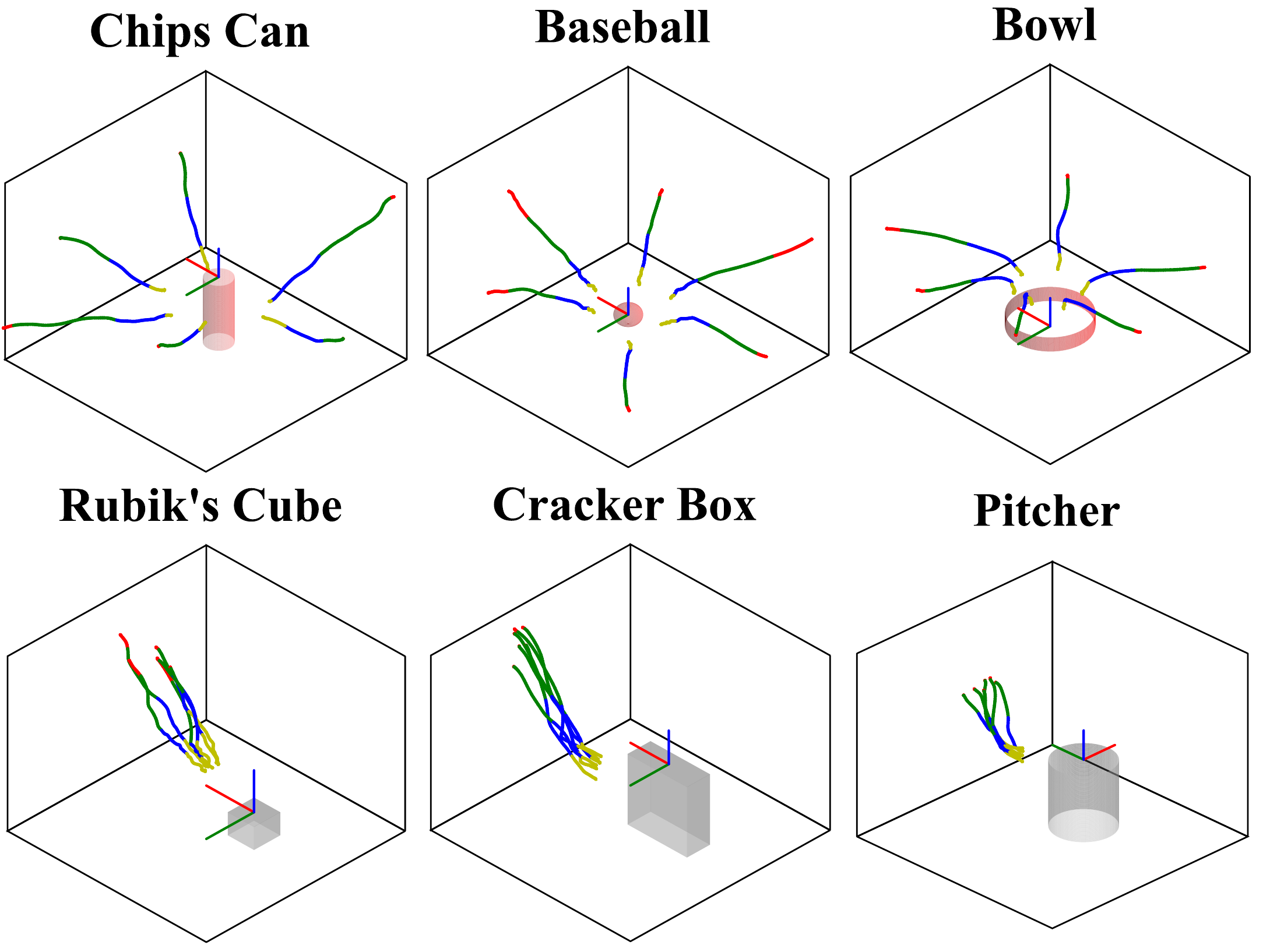}	
	\caption{Demonstrations for the grasping task with six different objects. The demonstrations are shown in the object frame and the trajectories with different colors are clustered with a GMM.}
	\label{Fig7}
\end{figure}

For the real robot experiment, we selected six different objects as targets, whose locations are tracked by using visual markers \cite{ARmarker}. A successful task is defined as the object being grasped firmly enough by the robot so that it can be held until the end of the motion. We demonstrated six grasping movements for each object with kinesthetic teaching, see \autoref{Fig7}. We fit a GMM with four components by considering the position in $\mathcal{M}_1$ augmented with timesteps information to cluster the demonstration into four phases (here, this number has been set by the experimenter). We then used the marginal distribution for the timesteps dimension as weights to compute Gaussian distributions for the four phases in $\mathcal{M}_1$, $\mathcal{M}_2$ and $\mathcal{M}_3$, which are used to form references for each manifold by weighted regression, see \autoref{Fig21} for an example of generated reference.

\begin{figure}
		\centering
 		\includegraphics[width=\columnwidth]{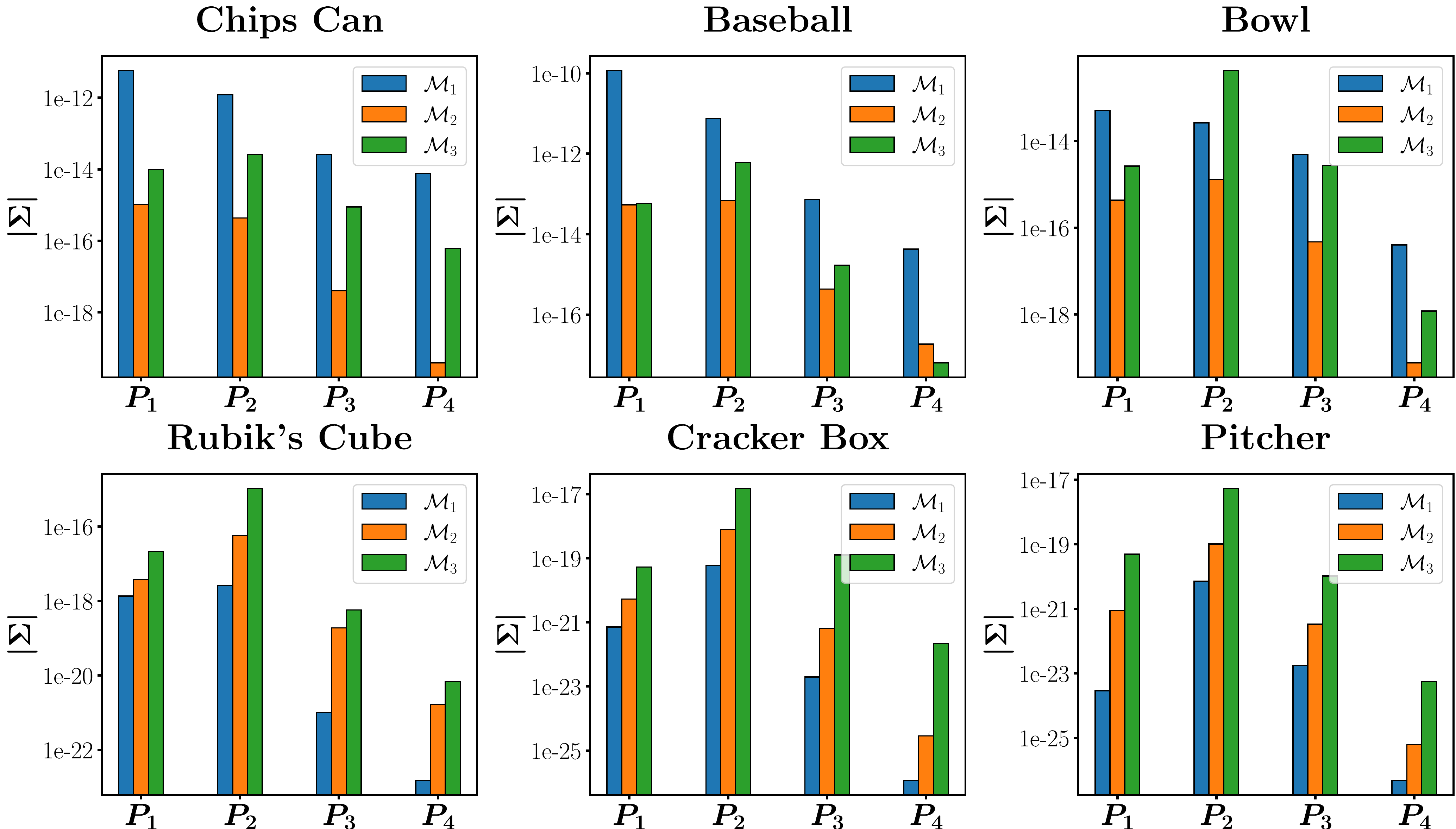}		
	\caption{Determinants of covariance matrices in the four phases (in logarithm scale). Blue, orange and green bars represent the values of $\mathcal{M}_1$, $\mathcal{M}_2$ and $\mathcal{M}_3$. The manifold with smallest determinant is the selected coordinate system for each phase of the movement.}
	\label{Fig15}
	\vspace{-0.3cm}
\end{figure}

\begin{figure}
	\centering
	\includegraphics[width=\columnwidth]{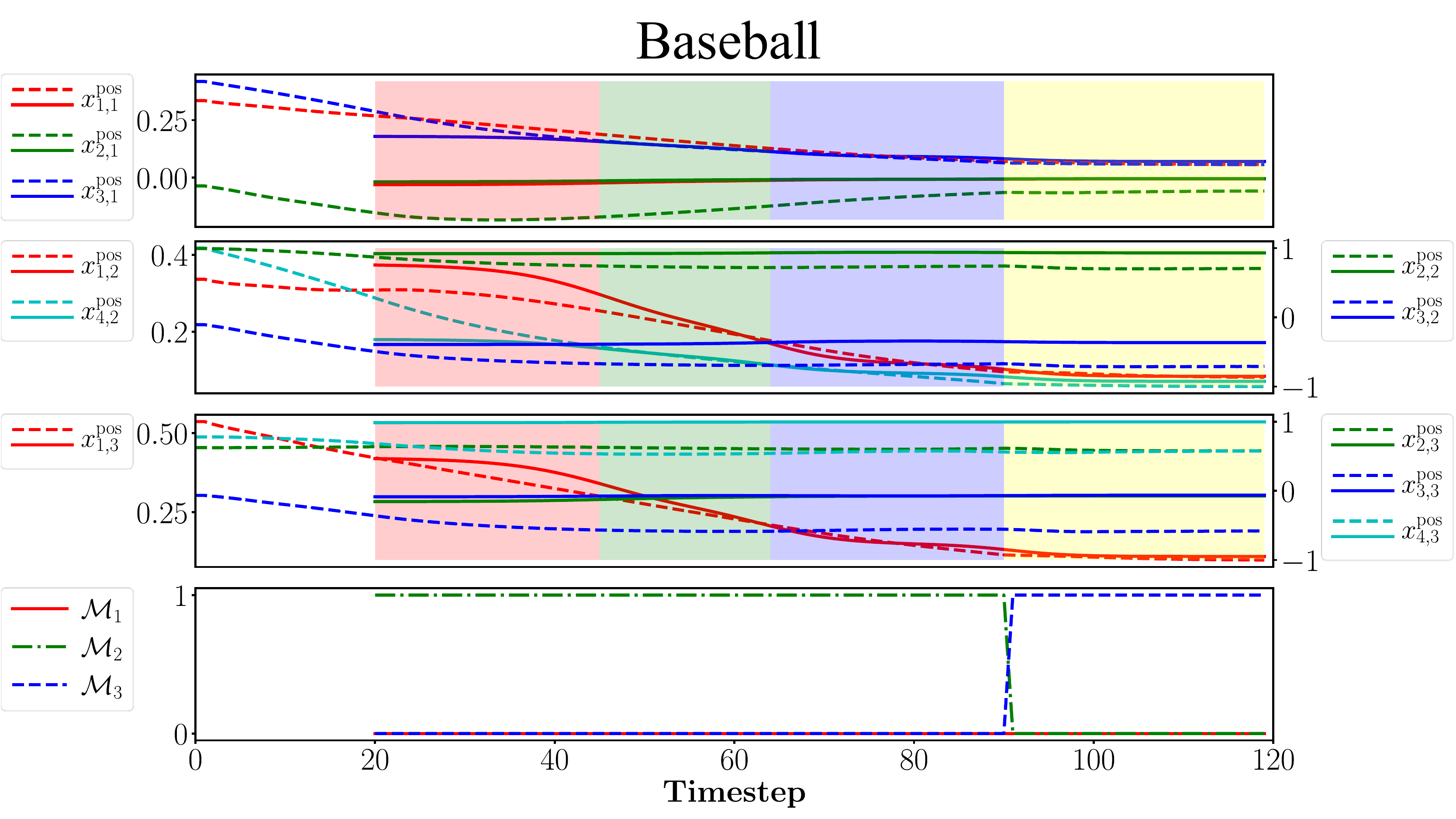}
	\caption{Timeline graphs. The solid lines represent the references in $\mathcal{M}_1$, $\mathcal{M}_2$ and $\mathcal{M}_3$ generated from a GMM with four components (four consecutive phases depicted in red, green, blue and yellow). The dashed lines represent a grasping movement generated from a different initial pose, by using the selected manifold at each time step. The bottom plot shows the selected manifold for the different timesteps (here, $\mathcal{M}_2$ then $\mathcal{M}_3$ are selected).}
	\label{Fig21}
	\vspace{-0.3cm}
\end{figure}

Coordinate systems are selected at each time step of the movements based on determinants of covariance matrices. These determinants are shown in \autoref{Fig15} for the different objects. We observe that the selected coordinate systems are related to the shapes of the objects and to the actions to perform with these object, which can also change during the task, see \autoref{Fig21} for an example.

For the chips can and bowl (with cylindrical shapes), $\mathcal{M}_2$ is selected by the algorithm to represent the motion. For prismatic objects (i.e. the Rubik's cube, the cracker box and pitcher handle), $\mathcal{M}_1$ is selected by the algorithm to represent the motion. For the baseball, although the determinants in $\mathcal{M}_2$ and $\mathcal{M}_3$ are close in the first three phases, $\mathcal{M}_2$ is selected for the first part of the movement, while $\mathcal{M}_3$ is selected for the last part of the movement. The statistical analysis of the demonstrations revealed that a cylindrical distribution was preferred at the beginning (when still far from the object), with a switch to a spherical coordinate system when the gripper is brought closer to the object, see \autoref{Fig21}.
This small preference of switching from a cylindrical geometry to a spherical geometry can be explained by the additional geometric factors coming into play, including the geometry of the robot (i.e., its reaching capability due to its kinematic chain), combined with the geometry of the gripper and the table on which the baseball is placed, favoring grasps from above, in order to avoid hitting the table with the gripper.

To evaluate the generalization capability of our approach, we conducted 50 reproduction trials for each object by varying the initial configurations of the robot randomly according to the Gaussian distribution representing the diversity of initial poses demonstrated by the user.

The optimal control problem is set so that the reference trajectories are active after 20 timesteps, allowing the robot to start from different initial configurations.

\begin{figure}
	\centering
	\includegraphics[width=\columnwidth]{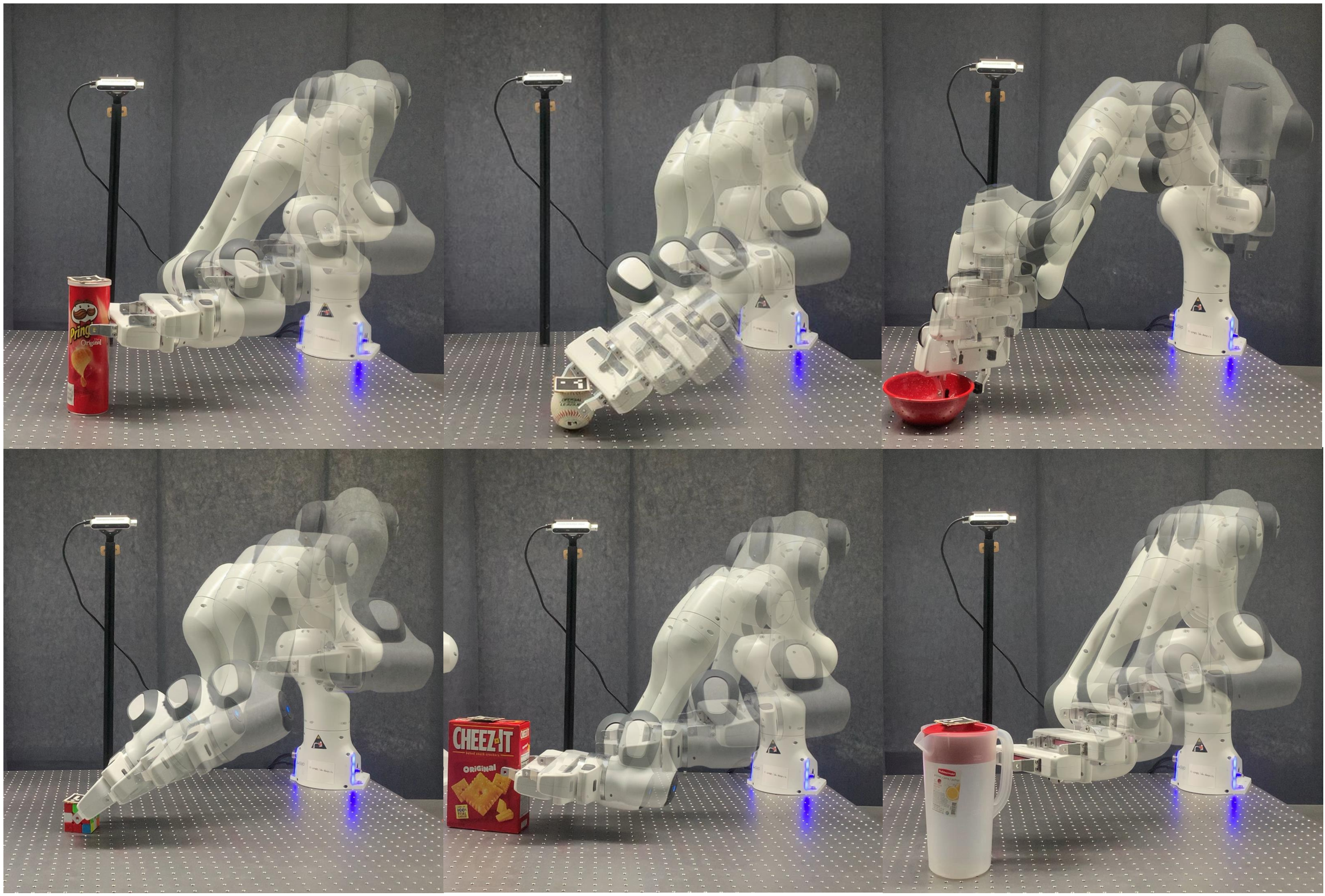}	
	\caption{Generalization trials for the grasping experiment.}
	\label{Fig12}
\end{figure}

\begin{figure}
	\centering
	\includegraphics[width=\columnwidth]{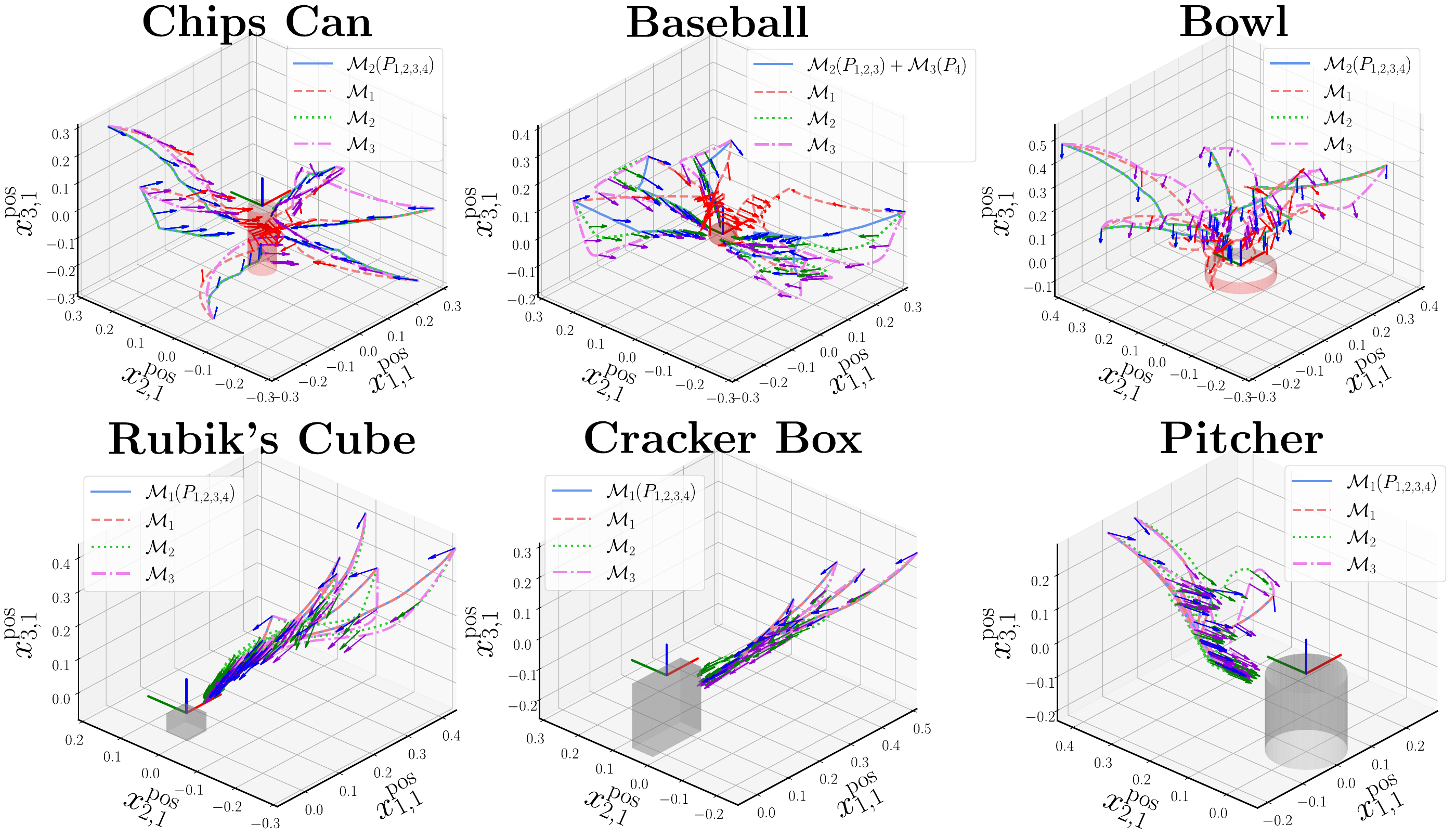}	
	\caption{Reproduced trajectories by using the optimal manifold method (blue solid lines) and by using only $\mathcal{M}_1$ (red dashed lines). The arrows represent the directions of the end-effector.}
	\label{Fig13}
	\vspace{-0.3cm}
\end{figure}

For each object, we compared the optimal manifold approach with the baseline of simply using $\mathcal{M}_1$ (i.e., with an encoding only based on a Cartesian coordinate system). The generated trajectories for these two methods are shown in \autoref{Fig13}. For the chips can, the bowl and the baseball, the pose of the movement generated in $\mathcal{M}_1$ is influenced by the orientation of the object, which causes the grasping motion to become clumsy. The results presented in \autoref{table1} involved 1200 different simulated reproduction trials, where the simulated setup matches the real robot setup. The attempts with less than 80\% success rate are marked in bold, in order to highlight the consistency of our approach, in contrast to the baselines considering a single coordinate system, which typically work for a subset of objects/actions but not for the other subset of objects/actions. Additional reproduction attempts on the real robot are presented in \autoref{Fig12} and in the accompanying video.
\begin{table}
	\renewcommand{\arraystretch}{1.2} 
	\setlength{\tabcolsep}{8pt}
	\centering 
	\caption{Success rates for grasping objects with different shapes.}
	\begin{tabular}{ccccc}
		\hline
		& $\mathcal{M}_1$ & $\mathcal{M}_{2}$ & $\mathcal{M}_{3}$ & Optimal $\mathcal{M}_{\hat{n}}$\\
		\hline
		Chips Can & \textbf{0/50} & 45/50 & 
		\textbf{30/50} & 42/50 ($\mathcal{M}_2$)\\
		Baseball & \textbf{2/50} & 41/50 & 45/50 & 44/50 ($\mathcal{M}_2$+$\mathcal{M}_3$)\\
		Bowl & \textbf{0/50} & 42/50 & \textbf{2/50} & 45/50 ($\mathcal{M}_2$) \\
		Rubik's Cube & 47/50 & \textbf{34/50} & 
		\textbf{39/50} & 46/50 ($\mathcal{M}_1$)\\
		Cracker Box & 45/50 & 43/50 & \textbf{26/50} & 47/50 ($\mathcal{M}_1$)\\
		Pitcher & 45/50 & 40/50 & \textbf{33/50} & 46/50 ($\mathcal{M}_1$)\\
		\hline
	\end{tabular} 
	\label{table1} 
\end{table}

\begin{figure}
	\centering
	\includegraphics[width=\columnwidth]{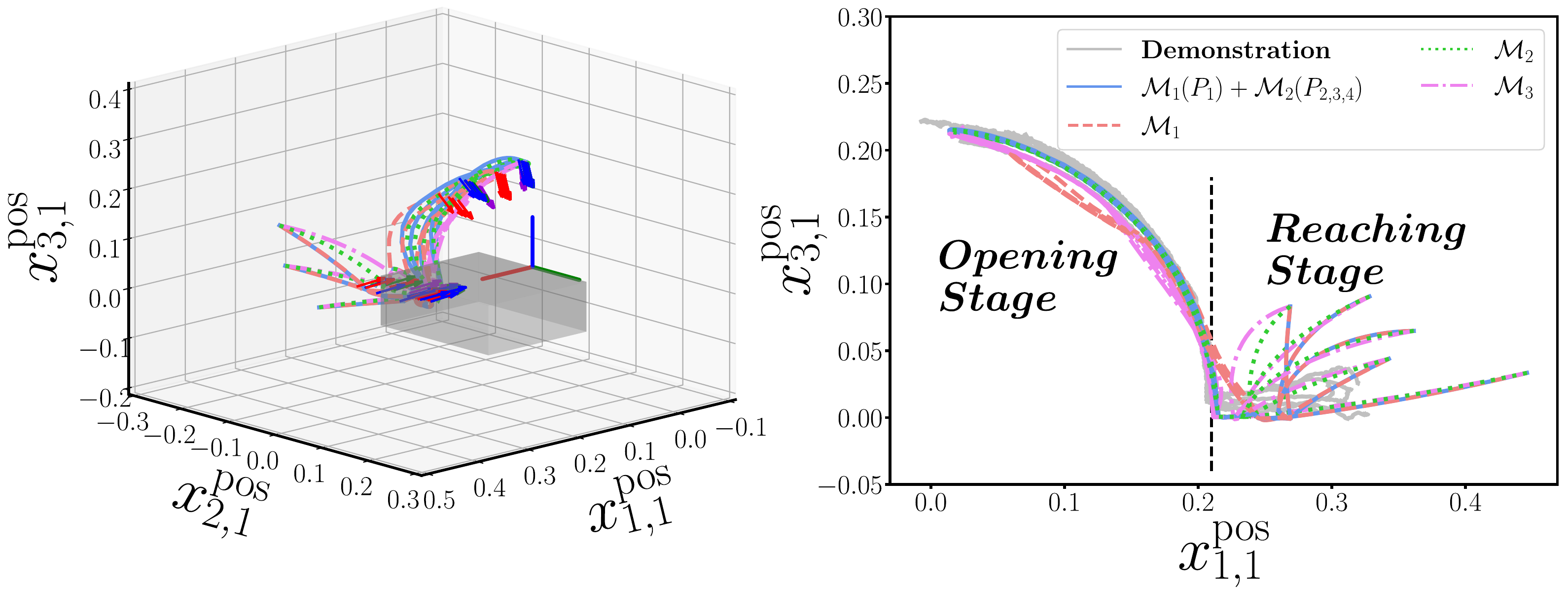}	
	\caption{Reproduced box opening movements for different initial states.}
	\label{Fig5}	
\end{figure}
\begin{figure}
	\centering
	\subfigure[Demonstrations]{
		\centering
		\includegraphics[width=0.4\columnwidth]{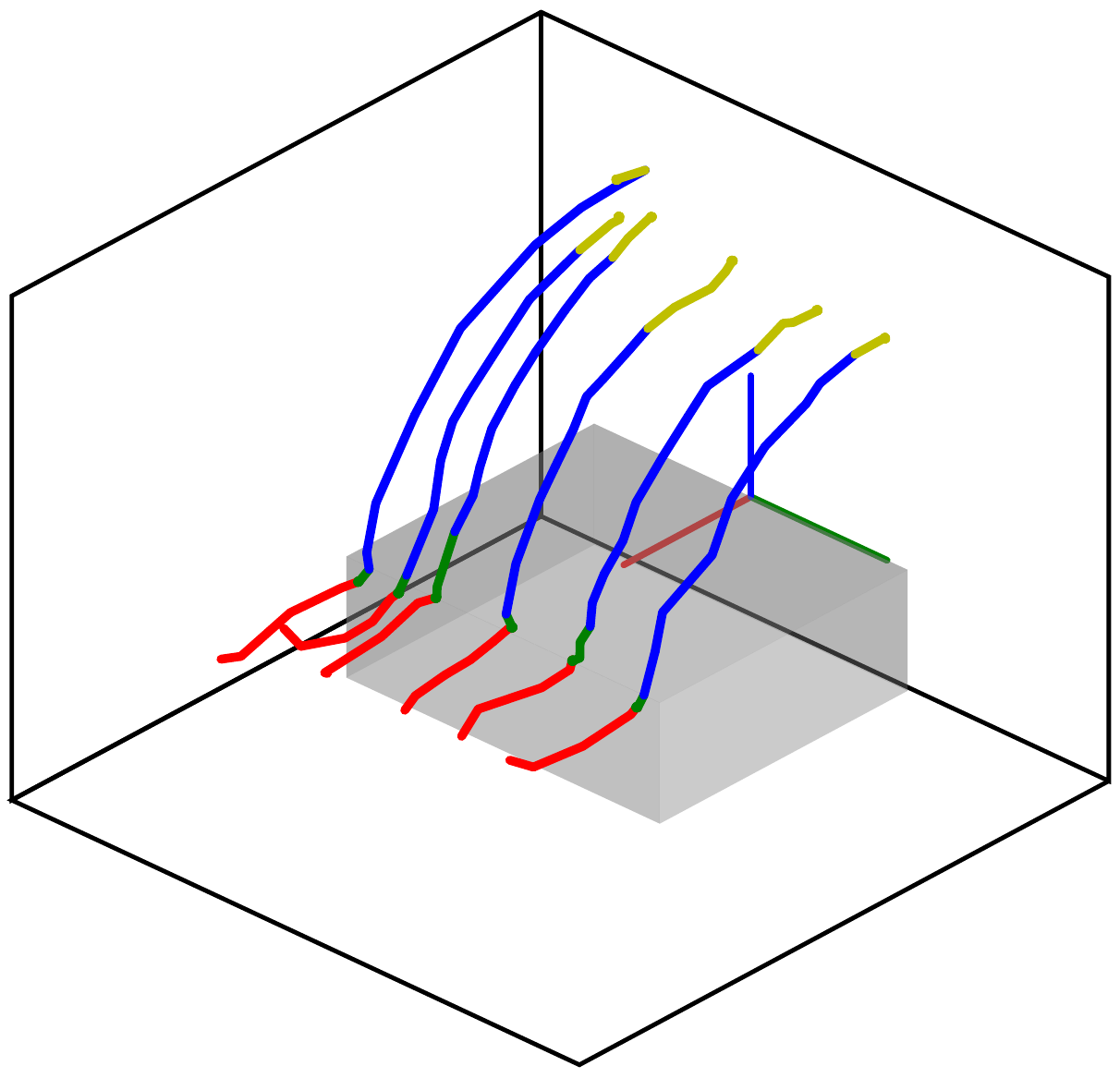}	
		\label{Fig4}	
	}
	\subfigure[Determinants of covariance matrices]{
		\includegraphics[width=0.45\columnwidth]{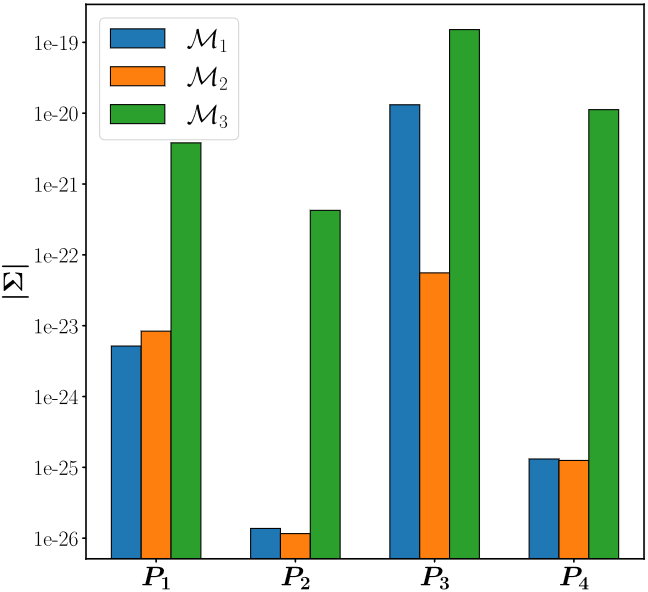} 
		\label{Fig16}
	}
	\subfigure[Timeline evolution]{
		\includegraphics[width=\columnwidth]{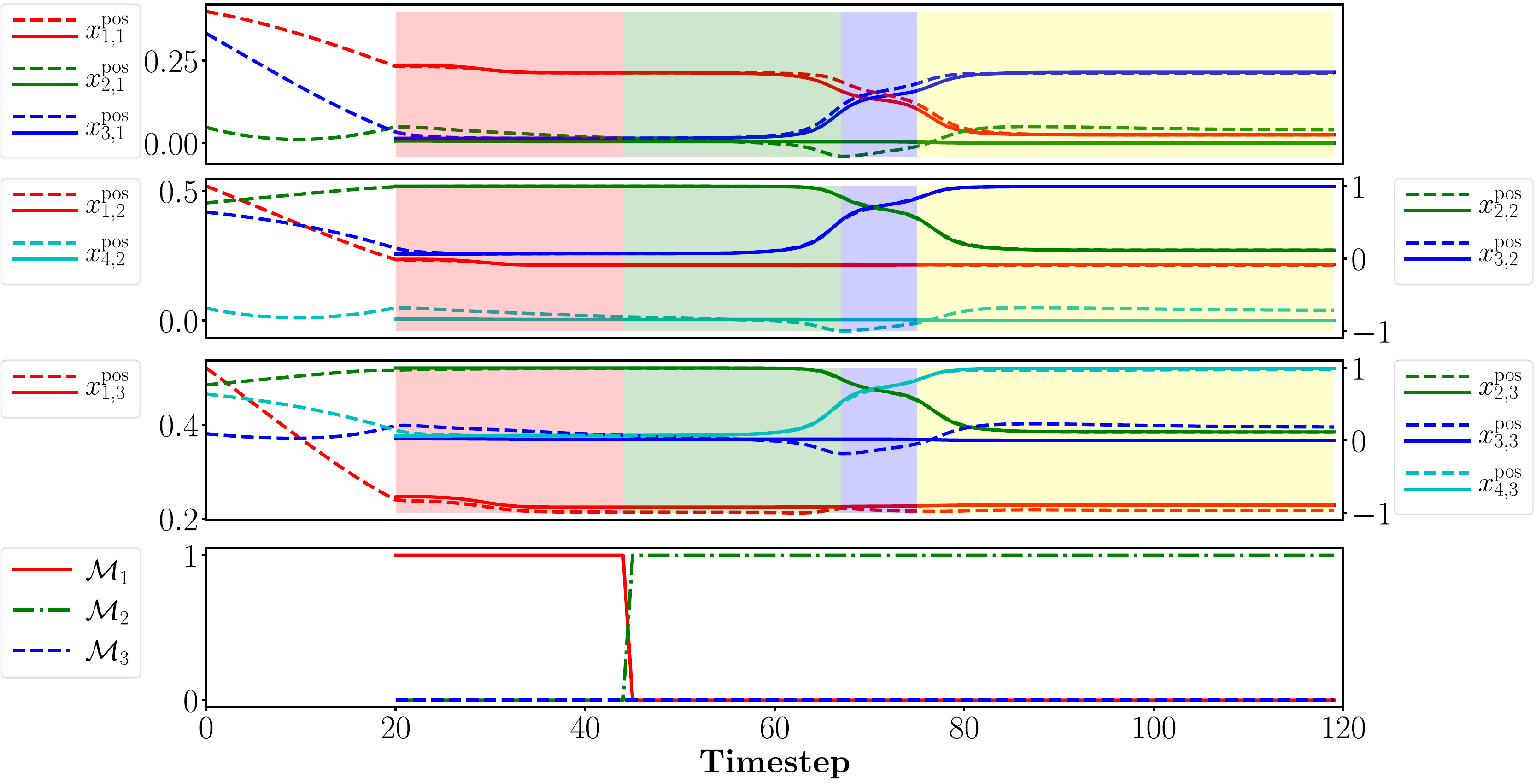} 
		\label{Fig20}
	}
	\caption{(a) Demonstrations of the box opening movement, with different colors representing different phases of the movement. (b) Determinants of covariance matrices in the four phases (in logarithmic scale). The manifold with smallest determinant is the selected coordinate system for each phase of the movement. (c) In the first three plots, the solid lines represent the references for $\mathcal{M}_1$, $\mathcal{M}_2$ and $\mathcal{M}_3$. The dashed lines represent a grasping movement generated from a different initial pose, by using the selected manifold at each time step. The bottom plot shows the selected manifold for the different timesteps (here, $\mathcal{M}_1$ then $\mathcal{M}_2$ are selected).}
	\label{FigBoxOpening}
\end{figure}

\begin{figure}
	\centering
	\includegraphics[width=\columnwidth]{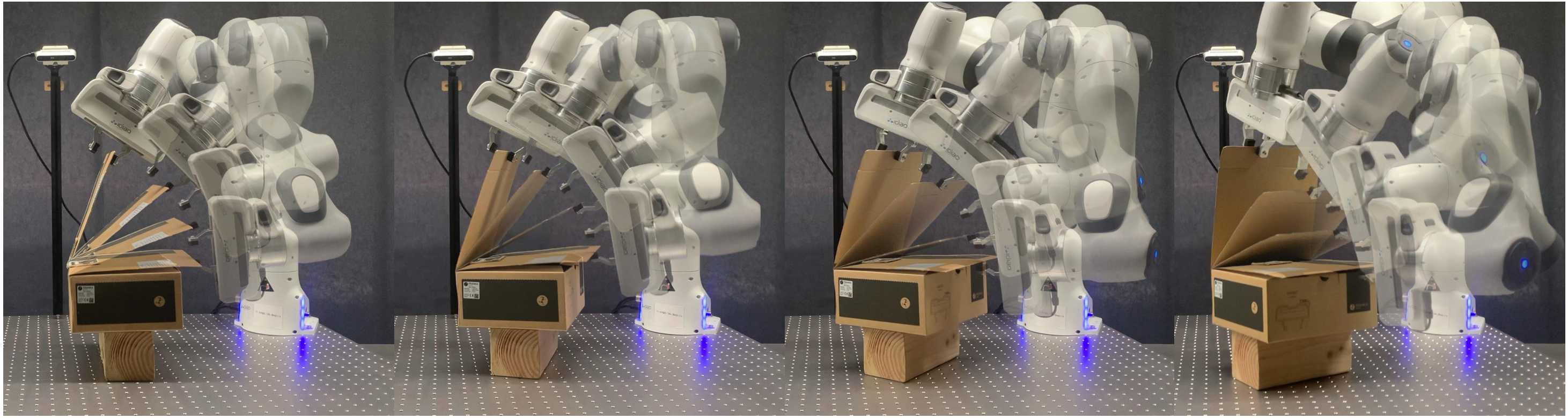}	
	\caption{Generalization trials for the box opening experiment.}
	\label{Fig14}
	\vspace{-0.3cm}
\end{figure}

\subsection{Box opening task with the Franka Emika robot}
We then tested our approach with a box opening task by using the same experiment protocol, with six demonstrations provided with a visual marker attached to the demonstrator's hand. A successful task is defined as the box being opened with a constant contact between the gripper and the box while opening (i.e., with the gripper not slipping and not applying an unnecessary excessive force on the lid). \autoref{FigBoxOpening} presents encoding and generalization results. The reproduced movements are shown in \autoref{Fig5}, with an example in \autoref{Fig14} (see also accompanying video), showing that the robot can successfully generalize the task by reaching and opening the box from different initial configurations of the box and the robot. In particular, we can see that the robot correctly learned to maintain a constant distance from the axis of rotation, which reflects the property of the opening motion. Extracting such property from demonstration is useful to open a box, but we also find similar affordances for objects and tools characterized by one or several rotation axes, such as turning the pages of a book, operating handles, valves or steering wheels, or opening doors, cabinets, dishwashers and fridges. By using only  $\mathcal{M}_1$ or $\mathcal{M}_3$, this property cannot be fulfilled, where the robot either cannot maintain the constant opening radius as in the demonstration or fails to reach, which can result in potential damage to the gripper or to the inability to open the box.


\section{Limitations of the current approach}
The approach presented here adopted a \emph{winner-takes-all} strategy by estimating at each timestep an optimal coordinate system to encode the task. Since our representation is probabilistic with full covariance matrices, another possible strategy would be to fuse multiple coordinate systems by relying on products of Gaussians, where coordinate systems with low variations in the demonstrations would have a higher impact on the movement generation than coordinate systems with high variations in the demonstrations, similarly to the approach adopted for task-parameterized models \cite{Calinon16JIST}, which also consider multiple coordinate systems, but only Cartesian frames of reference centered on different objects or landmarks. Treating the problem as a \emph{winner-takes-all} strategy can be viewed as a simplified approach, but its extension to a fusion problem is straightforward with the proposed probabilistic formulation. As it acts as some form of regularization, it has the potential of requiring fewer demonstrations, but it can only model skills requiring one manifold at a time (instead of fusion information from different manifolds), which further needs to be investigated. Moreover, due to the limitation of the timestep, if the initial state was far from the distribution of the demonstration such as prismatic object grasping, it would take more timesteps to reach the distribution range and continue the last movement within the limited timesteps, resulting in failure to reach the target. In the presented experiment, markers are used to track objects, which simplifies the localization of objects, but which also limits the applicability of the approach to real-world scenarios. Further work is required to combine the proposed manifold-based approach to more elaborated perception skills, including part-based detection of objects to provide local geometry information. The next step will also be to extend the current approach based on a movement representation of the task to force-aware trajectory generation.

\section{Conclusion}
In this paper, we proposed an approach considering different types of coordinate systems to reduce the number of demonstrations required to acquire object manipulation skills. We relied on an extension of Gaussian distribution on Riemannian manifolds to encode and detect invariant geometric features from demonstrations. By combining this representation with an optimal control strategy, we demonstrated that the robot was able to generate controllers for new situations by considering the symmetries and affordances of the object being manipulated and by exploiting the variations allowed by the task, resulting in a minimal intervention control strategy using various geometries. Our experimental results showed that our approach achieved higher generalization capability than the baseline of considering demonstrations only encoded in  Cartesian coordinate systems.



\bibliographystyle{ieeetr}
\bibliography{bib_iLQR}

\begin{thebibliography}{10}

\bibitem{Todorov02b}
E.~Todorov and M.~I. Jordan, ``A minimal intervention principle for coordinated
  movement,'' in {\em Advances in Neural Information Processing Systems
  ({NeurIPS})}, pp.~27--34, 2002.

\bibitem{Scholz99}
J.~P. Scholz and G.~Schoener, ``The uncontrolled manifold concept: identifying
  control variables for a functional task,'' {\em Experimental Brain Research},
  vol.~126, no.~3, pp.~289--306, 1999.

\bibitem{Wolpert11}
D.~M. Wolpert, J.~Diedrichsen, and J.~R. Flanagan, ``Principles of sensorimotor
  learning,'' {\em Nature Reviews}, vol.~12, pp.~739--751, 2011.

\bibitem{Sternad10}
D.~Sternad, S.-W. Park, H.~Mueller, and N.~Hogan, ``Coordinate dependence of
  variability analysis,'' {\em PLoS Comput. Biol.}, vol.~6, no.~4, pp.~1--16,
  2010.

\bibitem{Kelso09}
J.~A.~S. Kelso, ``Synergies: Atoms of brain and behavior,'' in {\em Progress in
  Motor Control} (D.~Sternad, ed.), vol.~629 of {\em Advances in Experimental
  Medicine and Biology}, pp.~83--91, Springer US, 2009.

\bibitem{Laumond18}
J.-P. Laumond, N.~Mansard, and J.-B. Lasserre, {\em Geometric and Numerical
  Foundations of Movements}.
\newblock Springer, 2018.

\bibitem{Bennequin09}
D.~Bennequin, R.~Fuchs, A.~Berthoz, and T.~Flash, ``Movement timing and
  invariance arise from several geometries,'' {\em PLoS Comput. Biol.}, vol.~5,
  no.~7, pp.~1--27, 2009.

\bibitem{Ganesh12}
G.~Ganesh and E.~Burdet, ``Motor planning explains human behaviour in tasks
  with multiple solutions,'' {\em Robotics and Autonomous Systems}, vol.~61,
  no.~4, pp.~362--368, 2013.

\bibitem{kaiser2019}
A.~Kaiser, J.~A. Ybanez~Zepeda, and T.~Boubekeur, ``A survey of simple
  geometric primitives detection methods for captured 3d data,'' in {\em
  Computer Graphics Forum}, vol.~38, pp.~167--196, Wiley Online Library, 2019.

\bibitem{romanengo2021}
C.~Romanengo, A.~Raffo, Y.~Qie, N.~Anwer, and B.~Falcidieno, ``Fit4cad: A point
  cloud benchmark for fitting simple geometric primitives in cad objects,''
  {\em Computers \& Graphics}, 2021.

\bibitem{li2019}
L.~Li, M.~Sung, A.~Dubrovina, L.~Yi, and L.~J. Guibas, ``Supervised fitting of
  geometric primitives to 3d point clouds,'' in {\em Proceedings of the
  IEEE/CVF Conference on Computer Vision and Pattern Recognition},
  pp.~2652--2660, 2019.

\bibitem{Mason81}
M.~T. Mason, ``Compliance and force control for computer controlled
  manipulators,'' {\em IEEE Trans.\ on Systems, Man, and Cybernetics}, vol.~11,
  no.~6, pp.~418--432, 1981.

\bibitem{borghesan2015}
G.~Borghesan, E.~Scioni, A.~Kheddar, and H.~Bruyninckx, ``Introducing geometric
  constraint expressions into robot constrained motion specification and
  control,'' {\em IEEE Robotics and Automation Letters}, vol.~1, no.~2,
  pp.~1140--1147, 2015.

\bibitem{Billard16chapter}
A.~G. Billard, S.~Calinon, and R.~Dillmann, ``Learning from humans,'' in {\em
  Handbook of Robotics} (B.~Siciliano and O.~Khatib, eds.), ch.~74,
  pp.~1995--2014, Secaucus, NJ, USA: Springer, 2016.
\newblock 2nd Edition.

\bibitem{Calinon19entry}
S.~Calinon, ``Learning from demonstration (programming by demonstration),'' in
  {\em Encyclopedia of Robotics} (M.~H. Ang, O.~Khatib, and B.~Siciliano,
  eds.), Springer, 2019.

\bibitem{PerezDArpino17}
C.~P\'erez-D'Arpino and J.~A. Shah, ``C-learn: Learning geometric constraints
  from demonstrations for multi-step manipulation in shared autonomy,'' in {\em
  Proc.\ {IEEE} Intl Conf.\ on Robotics and Automation ({ICRA})},
  pp.~4058--4065, 2017.

\bibitem{Subramani18}
G.~Subramani, M.~Zinn, and M.~Gleicher, ``Inferring geometric constraints in
  human demonstrations,'' in {\em Proc.\ Conference on Robot Learning
  ({CoRL})}, pp.~223--236, 2018.

\bibitem{Calinon14ICRA}
S.~Calinon, D.~Bruno, and D.~G. Caldwell, ``A task-parameterized probabilistic
  model with minimal intervention control,'' in {\em Proc.\ {IEEE} Intl Conf.\
  on Robotics and Automation ({ICRA})}, (Hong Kong, China), pp.~3339--3344,
  2014.

\bibitem{Mayne66}
D.~Mayne, ``A second-order gradient method for determining optimal trajectories
  of non-linear discrete-time systems,'' {\em International Journal of
  Control}, vol.~3, no.~1, pp.~85--95, 1966.

\bibitem{Li04}
W.~Li and E.~Todorov, ``Iterative linear quadratic regulator design for
  nonlinear biological movement systems,'' in {\em Proc.\ Intl Conf.\ on
  Informatics in Control, Automation and Robotics (ICINCO)}, pp.~222--229,
  2004.

\bibitem{Calinon16JIST}
S.~Calinon, ``A tutorial on task-parameterized movement learning and
  retrieval,'' {\em Intelligent Service Robotics}, vol.~9, pp.~1--29, January
  2016.

\bibitem{Eppner13}
C.~Eppner and O.~Brock, ``Grasping unknown objects by exploiting shape
  adaptability and environmental constraints,'' in {\em Proc.\ {IEEE/RSJ} Intl
  Conf.\ on Intelligent Robots and Systems ({IROS})}, pp.~4000--4006, 2013.

\bibitem{Calinon20RAM}
S.~Calinon, ``Gaussians on {R}iemannian manifolds: Applications for robot
  learning and adaptive control,'' {\em {IEEE} Robotics and Automation Magazine
  ({RAM})}, vol.~27, pp.~33--45, June 2020.

\bibitem{Lembono21IROS}
T.~S. Lembono and S.~Calinon, ``Probabilistic iterative {LQR} for short time
  horizon {MPC},'' in {\em Proc.\ {IEEE/RSJ} Intl Conf.\ on Intelligent Robots
  and Systems ({IROS})}, pp.~556--562, 2021.

\bibitem{Pennec06}
X.~Pennec, ``Intrinsic statistics on {R}iemannian manifolds: Basic tools for
  geometric measurements,'' {\em Journal of Mathematical Imaging and Vision},
  vol.~25, no.~1, pp.~127--154, 2006.

\bibitem{SimoSerra16}
E.~Simo-Serra, C.~Torras, and F.~Moreno-Noguer, ``3{D} human pose tracking
  priors using geodesic mixture models,'' {\em International Journal of
  Computer Vision}, vol.~122, no.~2, pp.~388--408, 2017.

\bibitem{Zeestraten17RAL}
M.~J.~A. Zeestraten, I.~Havoutis, J.~Silv\'erio, S.~Calinon, and D.~G.
  Caldwell, ``An approach for imitation learning on {R}iemannian manifolds,''
  {\em {IEEE} Robotics and Automation Letters ({RA-L})}, vol.~2,
  pp.~1240--1247, June 2017.

\bibitem{ARmarker}
S.~Garrido-Jurado, R.~Muñoz-Salinas, F.~Madrid-Cuevas, and M.~Marín-Jiménez,
  ``Automatic generation and detection of highly reliable fiducial markers
  under occlusion,'' {\em Pattern Recognition}, vol.~47, no.~6, pp.~2280--2292,
  2014.

\end{thebibliography}

\end{document}